\documentclass[journal]{IEEEtran}

\newcommand{\qed}{\nobreak \ifvmode \relax \else
      \ifdim\lastskip<1.5em \hskip-\lastskip
      \hskip1.5em plus0em minus0.5em \fi \nobreak
      \vrule height0.75em width0.5em depth0.25em\fi}

\pdfoutput=1
\usepackage{etoolbox}

\usepackage[cmex10]{amsmath}
\usepackage{amssymb}
\usepackage{notacion2}
\usepackage[normalem]{ulem}

\usepackage{enumitem}

\usepackage{upgreek}

\ifx\pdfoutput\undefined
\usepackage{graphicx}
\else
\usepackage[pdftex]{graphicx}
\usepackage{epstopdf}
\fi
\usepackage[svgnames, x11names]{xcolor}

\usepackage{environ}

  \usepackage[caption=false,font=normalsize,labelfont=sf,textfont=sf]{subfig}

\begin{document}
%
\title{The Generalized Complex Kernel Least-Mean-Square Algorithm}
%
%
%

\author{Rafael~Boloix-Tortosa*,~\IEEEmembership{Member,~IEEE,}
       Juan~Jos\'e~Murillo-Fuentes,~\IEEEmembership{Senior~Member,~IEEE,}
        Sotirios A. Tsaftaris,~\IEEEmembership{Senior~Member,~IEEE}
\thanks{R. Boloix-Tortosa, 
        and J.J. Murillo-Fuentes are with the Dep. de Teor\'ia de la Se\~nal y Comunicaciones, Escuela T\'ecnica Superior de Ingenier\'ia, Universidad de Sevilla, Camino de los Descubrimientos sn, 41092 Sevilla, Spain. e-mail: \{rboloix,murillo\}@us.es.}
\thanks{S. A. Tsaftaris is with the School of Engineering, The University of Edinburgh, Edinburgh EH8 3FB, U.K., and also with The Alan Turing Institute, London NW1 2DB, U.K.}
\thanks{Thanks to the Spanish Government (Ministerio de Econom\'ia y Competitividad, TEC2016-78434-C3-02-R, and Ministerio de Educaci\'on, Cultura y Deporte, Subprograma Estatal de Movilidad (PRX18/00523), del Plan Estatal de I+D+I) and European Union (FEDER) for funding.
}
}

\hyphenation{op-tical net-works semi-conduc-tor hy-per-pa-ra-me-ters ge-ne-ra-tion ge-ne-ra-ted}

\maketitle

\begin{abstract}
  
We propose a novel adaptive kernel based regression method for complex-valued signals: the generalized complex-valued kernel least-mean-square (gCKLMS). We borrow from the new results on widely linear reproducing kernel Hilbert space (WL-RKHS) for nonlinear regression and complex-valued signals, recently proposed by the authors. This paper shows that in the adaptive version of the kernel regression for complex-valued signals we need to include another kernel term, the so-called pseudo-kernel. This new solution is endowed with better representation capabilities in complex-valued fields, since it can efficiently decouple the learning of the real and the imaginary part. Also, we review previous realizations of the complex KLMS algorithm and its augmented version to prove that they can be rewritten as particular cases of the gCKLMS. Furthermore, important conclusions on the kernels design are drawn that help to greatly improve the convergence of the algorithms. In the experiments, we revisit the nonlinear channel equalization problem to highlight the better convergence of the gCKLMS compared to previous solutions. Also, the flexibility of the proposed generalized approach is tested in a second experiment with non-independent real and imaginary parts. The results illustrate the significant performance improvements of the gCKLMS approach when the complex-valued signals have different properties for the real and imaginary parts. 

 \end{abstract}

\begin{IEEEkeywords}
LMS, complex-valued, RKHS, kernel methods.
\end{IEEEkeywords}

%
\IEEEpeerreviewmaketitle

\setlength{\arraycolsep}{0.1em} 

\section{Introduction}

\IEEEPARstart{C}{omplex-valued} signals model many systems in diverse applications such as electromagnetism, telecommunications, optics or acoustics, among others. Complex-valued signal processing is thus of fundamental interest as it provides a natural way to represent some signals and transformations involved in those systems. 
While the linear case has been widely studied (see for example \cite{Schreier06} and references therein), nonlinear processing still remains an open problem. Nonlinear processing of complex-valued signals has been tackled, among others, from the point of view of neural networks \cite{hirose13}, \cite{Valle14}, nonlinear adaptive filtering \cite{Mandic09}, or reproducing kernel Hilbert spaces (RKHS) \cite{Scholkopf02}. This latter field is gaining increasing interest within the signal processing community as it provides a simple but elegant way to treat nonlinearities. Complex kernel-based algorithms have been lately proposed for regression \cite{Bouboulis11,Vaerenbergh12,Boloix18}, kernel principal component analysis \cite{Papaioannou14} or classification \cite{Steinwart06}.

Regarding complex-valued regression within the RKHS framework, we have recently highlighted in \cite{Boloix17} the need of a new term: the \emph{pseudo-kernel}. We redefined the kernel based regularized least squares regression to include the pseudo-kernel, and the resulting structure resembles that of the widely linear (WL) solutions, being capable of learning any complex-valued function effectively. As discussed in \cite{Boloix17}, the need for a pseudo-kernel can be justified in cases where the real and imaginary parts are correlated and learning them independently is, at best, suboptimal. Also, a pseudo-kernel is needed when the real and imaginary parts are not best represented by the same kernel, i.e., the same measure of similarity. Furthermore, we analyzed in \cite{Boloix17} the structure of the kernel and pseudo-kernel, and discussed how to design these functions, and when should they be real or complex-valued. As a result, two important remarks were made. First, if the real and imaginary parts of the output are independent, then the kernel and pseudo-kernel should be real-valued. Second, if the real and imaginary parts of the output have different properties in terms of similarity, the pseudo-kernel is needed. On the contrary, the pseudo-kernel vanishes if the real and imaginary parts of the output are independent but have same properties in terms of similarity, i.e., the same kernel can be used for the real and imaginary parts. 

In the design of adaptive nonlinear approaches, the authors in \cite{Bouboulis11} address the problem of adaptive filtering of complex signals and calculate the gradient of cost functions by using Wirtinger's derivatives. Two alternatives are described. The first alternative proposes using real kernels, by means of the technique called \emph{complexification} of real RKHSs. The second one proposes the use of complex kernels, in particular, the complex Gaussian kernel \cite{Steinwart06}. By means of these two alternatives they develop two realizations of the kernel least-mean-square (KLMS) algorithm \cite{Liu08}. The same complex Gaussian kernel is also adopted in \cite{OgunfunmiP11} and in \cite{Bouboulis12}, where they propose to introduce WL adaptive filters in complex RKHS to solve a nonlinear filtering task. Augmented or WL filters consider both the original values of the signal data and their conjugates \cite{Picinbono95}, and are able to capture the full second-order statistical characteristics of the signal \cite{Picinbono97,Schreier03}. This is highlighted in \cite{Bouboulis12} as a key starting point to develop the augmented complex KLMS. The authors remark that ``the natural choice for kernels, in the context of the WL filtering structure, are the pure complex kernels". Other augmented complex kernel algorithms have also been proposed \cite{Kuh09,Kuh10,Tobar12}. 

In this paper, we propose a novel generalized formulation for the adaptive complex KLMS algorithm. In light of the findings in \cite{Boloix17}, we herein develop the generalized complex KLMS algorithm, that we call gCKLMS, which includes a kernel and a pseudo-kernel term. We show that the kernel and pseudo-kernel in the gCKLSM have the same structures found in \cite{Boloix17}, and we can use the analysis in that work to design these two functions. Unlike in \cite{Bouboulis12}, we conclude that a complex kernel is not always the best choice for the adaptive complex KLMS algorithm, as we will show later in the experiments in \SEC{Exp}. We also show that previous proposed complex or augmented complex KLMS algorithms are limited, as they are just particular simplifications of the more general formulation proposed in this paper.

Our starting point is the definition of the RKHS for the \emph{composite} representation of complex-valued functions. This is the representation of a complex-valued function as a two-dimensional real-valued vector function, obtained by stacking the real part of the function over the imaginary part. We devote \SEC{VectorRKHS} to review the theory of kernels for multi-task learning \cite{Micchelli05} as a suitable feature map representation of the composite function. In \SEC{CompoKLMS} we develop the KLMS algorithm for the composite representation. The composite representation of a complex-valued function is related to its \emph{augmented} representation \cite{Schreier03}. This is the representation as a two dimensional complex vector with the complex-valued function on top of its complex conjugate. By using this relationship, the formulation for the gCKLMS algorithm is found in \SEC{AKLM}. In this section we also show the equations of the kernel and pseudo-kernel terms. In \SEC{previ} we compare the gCKLMS with other complex KLMS algorithms in the literature to show that they are particular cases of the gCKLMS. Experiments are included in \SEC{Exp}, where the gCKLMS algorithm is tested first in the context of a nonlinear channel equalization task, and then in the learning of samples of a filtered random process. These experiments show that the gCKLMS outperforms other KLMS algorithms, as it has both a kernel and a pseudo-kernel term. By making use of the remarks in \cite{Boloix17}, we have more suitable designs for the kernel and the pseudo-kernel that greatly improve the predictions. We end the paper with some conclusions in \SEC{Conclu}.

In the notation used throughout the paper, bold lower-case letters are used to denote vectors, while matrices are denoted using bold upper-case letters. 
For matrix $\vect{A}$, $\entry{\vect{A}}{l}{q}$ is its $(l,q)$ entry. To denote the $i$-th sample of a vector or signal we use, respectively, $\vect{a}(i)$ and $a(i)$. $\Re\left\{a\right\}$ is the real part of $a$. Transpose operation is represented by $\trs$, while $\her$ represents the Hermitian and $\cnj$ complex conjugation. $\E[\cdot]$ is the expectation operator.  
\section{RKHS of composite vector-valued functions} \LABSEC{VectorRKHS}

A complex function $\f(\x)=\f\rr(\x)+\j\f\jj(\x)$ can be represented as a composite vector-valued function $\fv\com{(\x)}=[\f\rr(\x)\; \f\jj(\x)]\trs\in \RN^{2}$, also known as the dual real channel (DRC) formulation, by stacking its real part on its imaginary part. The definition of the RKHS for vector-valued functions \cite{Micchelli05} parallels the one for scalar functions \cite{Aronszajn50}, with the main difference that the reproducing kernel is now matrix-valued \cite{Alvarez12}, \cite{Micchelli05}. 

Let $\mathcal{H}$ be a Hilbert space of functions $\fv$ on a set $\mathcal{X}$ with values in $\mathcal{Y}$. $\mathcal{H}$ is a RKHS when for any $\x\in\mathcal{X}$ and any $\yv\in\mathcal{Y}$ the linear functional which maps $\fv$ to $(\yv,\fv(\x))_\mathcal{Y}$ is continuous on $\mathcal{H}$ \cite{Micchelli05}. Here, $(\cdot,\cdot)_\mathcal{Y}$ represents the inner product in the Hilbert space $\mathcal{Y}$, while $\langle\cdot,\cdot\rangle_\mathcal{H}$ is the inner product in $\mathcal{H}$. 

From the Riesz Lemma, for every $\x\in\mathcal{X}$ and  $\yv\in\mathcal{Y}$ there is a linear operator $\K_{\x}: \mathcal{Y}\rightarrow\mathcal{H}$, such that $(\yv,\fv(\x))_\mathcal{Y}=\langle \K_{\x}\yv,\fv\rangle_\mathcal{H}$. Let us now introduce the linear operator $\K(\x,\x\new): \mathcal{Y}\rightarrow\mathcal{Y}$, for every $\x,\x\new\in\mathcal{X}$, defined by $\K(\x,\x\new)\yv:=(\K_{\x\new}\yv)(\x)$. 

We say that $\K:\mathcal{X}\times\mathcal{X}\rightarrow\mathcal{L(Y)}$, where $\mathcal{L(Y)}$ denotes the set of all bounded linear operators from $\mathcal{Y}$ to itself, is a matrix-valued kernel \cite{Micchelli05} (or operator-valued kernel if $\mathcal{Y}$ is not finite dimensional \cite{Caponnetto08}) if it satisfies the following properties for every $\x,\x\new\in\mathcal{X}$:
\begin{enumerate}[label=(\emph{\alph*})]
\item For every $\yv,\yv\new\in\mathcal{Y}$, we have $(\yv,\K(\x,\x\new)\yv\new)_\mathcal{Y}=\langle \K_{\x}\yv,\K_{\x\new}\yv\new\rangle_\mathcal{H}$.
\item $\K(\x,\x\new)=\bar{\K}(\x\new,\x)$, and $\K(\x,\x)\in\mathcal{L_{+}(Y)}$, where $\bar{\K}$ denotes the adjoint and $\mathcal{L_{+}(Y)}$ the set of all positive semi-definite bounded linear operators, i.e., $(\yv,\K(\x,\x)\yv)_\mathcal{Y}\ge0$ for any $\yv\in\mathcal{Y}$.
\item For any positive integer $m$, we have that $\sum_{l,q\in\{1,\cdots,m\}}(\yv_{q},\K(\x_{q},\x_{l})\yv_{l})_\mathcal{Y}\ge0$, for any $\x_{l},\x_{q}\in\mathcal{X}$, $\yv_{l},\yv_{q}\in\mathcal{Y}$.
\end{enumerate}
Proof of these properties can be found in \cite{Micchelli05}. Also, it can be shown that if $\K$ is a kernel then there exists a unique (up to an isometry) RKHS of
functions from $\mathcal{X}$ to $\mathcal{Y}$ which admits $\K$ as the reproducing kernel.

In the case of $\mathcal{Y}=\RN^{2}$, the kernel function $\K$ takes values as $2\times 2$ matrices and, from property (\emph{a}), the matrix elements can be found as:
\begin{align}
\entry{\K(\x,\x\new)}{l}{q}=\langle \K_{\x}\vect{e}_l,\K_{\x\new}\vect{e}_q\rangle_\mathcal{H},
\end{align}
where $\vect{e}_l,\vect{e}_q$ are the standard coordinate bases in $\RN^{2}$, for ${l,q\in\{1,2\}}$. 

\subsection{Feature map}

We next define a suitable feature map representation for the matrix-valued kernel that will be later useful in deriving the gCKLMS algorithm.

Every kernel $\K$ admits a feature map representation. A feature map is a continuous function $\boldsymbol{\upphi}: \mathcal{X}\rightarrow\mathcal{L(Y,W)}$, where $\mathcal{L(Y,W)}$ denotes all bounded linear operators from $\mathcal{Y}$ into the feature Hilbert space $\mathcal{W}$ \cite{Caponnetto08}. If $\bar{\boldsymbol{\upphi}}(\x)$ is the adjoint of ${\boldsymbol{\upphi}}(\x)$, it is in $\mathcal{L(W,Y)}$, and
\begin{align}
\K(\x,\x\new)=\bar{\boldsymbol{\upphi}}(\x)\boldsymbol{\upphi}(\x\new),
\end{align}
for any $\x,\x'\in\mathcal{X}$.

In the case of finite dimensional Hilbert spaces $\mathcal{Y}=\RN^{2}$ and $\mathcal{W}=\RN^{m}$, relative to standard basis of both spaces $\boldsymbol{\upphi}(\x)$ is a $m\times 2$ matrix. Each entry of this matrix, $\entry{\boldsymbol{\upphi}(\x)}{p}{q}=\phi_{pq}(\x)$ is a scalar-valued continuous function of $\x\in\mathcal{X}$, and each entry of the kernel is 
\begin{align}
\entry{\K(\x,\x\new)}{l}{q}=\sum_{i\in\{1,\cdots,m\}}\phi_{il}(\x)\phi_{iq}(\x\new).
\end{align}
Note that when $\mathcal{Y}=\RN$, then ${\boldsymbol{\upphi}}(\x)\in\mathcal{W}$, but this is not the case here. 

\section{The composite KLMS algorithm} \LABSEC{CompoKLMS}

Consider the training sequence of input-output pairs $\{(\x(1),\y(1)), . . . ,(\x(N),\y(N))\}$ where $\y(n)\in\CN$ and $\x(n)\in\CN^d$. The goal is to uncover the underlying complex-valued function $f(\x(i))$ based on these examples, so that to
minimize the mean square error $J=\E[|\y(i)-f(\x(i))|^2]=\E[|e(i)|^2]$. By using the composite notation, this can be written as 
\begin{align}
J&=\E[|e(i)|^2]=\E[\left( {\y\rr}{(i)}-\f\rr(\x(i))\right)^2+\left({\y\jj}{(i)}-\f\jj(\x(i))\right)^2]\nonumber\\
&=\E[\left(\yv\com(i)-\fv\com(\x(i))\right)\trs\left(\yv\com(i)-\fv\com(\x(i))\right)],
\end{align}
where $\fv\com{(\x)}=[\f\rr(\x)\; \f\jj(\x)]\trs$ and $\yv\com=[\y\rr \; \y\jj]\trs$.

The least-mean-square (LMS) algorithm would consider a linear input-output mapping, i.e., $f(\x(i))=\vect{w}\her\x(i)$, and compute the weight vector $\vect{w}$ adaptively using stochastic gradient descent updates \cite{Widrow75}. 
However, instead of a direct linear input-output mapping, the KLMS \cite{Liu08} is performed on the transformed inputs by using the feature map. We propose here to use the composite notations and the theory for RKHS of composite vector-valued functions described in the previous section. Therefore, we use the feature map $\boldsymbol{\upphi}: \mathcal{X}\rightarrow\mathcal{L(Y,W)}$ and set  $\fv\com(\x)=\bar{\boldsymbol{\upphi}}(\x)\vect{w}$, where $\vect{w}\in\mathcal{W}$. 

Note that in the general case $\mathcal{W}$ could be an infinite dimensional Hilbert space. For the particular case of $\mathcal{W}=\RN^{m}$, since $\mathcal{Y}=\RN^{2}$ then ${\boldsymbol{\upphi}}(\x)=[\PHI\rr(\x)\; \PHI\jj(\x)]$ is an $m\times 2$ matrix, where $\PHI\rr(\x)$ and $\PHI\jj(\x)$ are its first and second column, respectively, and $\bar{\boldsymbol{\upphi}}(\x)={\boldsymbol{\upphi}}\trs(\x)$:
\begin{align}
\fv\com(\x)&=\begin{bmatrix}\f\rr(\x)\\ \f\jj(\x)\end{bmatrix}={\boldsymbol{\upphi}}\trs(\x)\vect{w}=\begin{bmatrix}\PHI\trs\rr(\x)\\\PHI\trs\jj(\x)\end{bmatrix}\vect{w}.
\end{align}

The objective is now the minimization of
\begin{align}
J(\vect{w})=\E\left[\left(\yv\com(i)-{\boldsymbol{\upphi}}\trs\hspace{-1mm}(\x(i))\vect{w}\right)\trs\hspace{-1mm}\left(\yv\com(i)-{\boldsymbol{\upphi}}\trs\hspace{-1mm}(\x(i))\vect{w}\right)\right].
\end{align}
It is easy to show that the gradient is 
\begin{align}
\pder{J(\vect{w})}{\vect{w}}&=-2\E\left[{\boldsymbol{\upphi}}(\x(i))\left(\yv\com(i)-{\boldsymbol{\upphi}}\trs(\x(i))\vect{w}\right)\right]\nonumber\\
&=-2\E[{\boldsymbol{\upphi}}(\x(i))\matr{e}\com(i)],
\end{align}
and the update equation for $\matr{w}$ using the stochastic gradient yields
\begin{align}\LABEQ{updatew}
\matr{w}(i)=\matr{w}(i-1)+2\mu{\boldsymbol{\upphi}}(\x(i))\matr{e}\com(i).
\end{align}
If we set $\matr{w}(0)=\matr{0}$, the repeated application of the weight-update equation \EQ{updatew} yields
\begin{align}
\matr{w}(i)=2\mu\sum_{l=1}^{i}{\boldsymbol{\upphi}}(\x(l))\matr{e}\com(l).
\end{align}
At instant $i$ the output can be estimated using the last updated weights, $\matr{w}(i-1)$, as $\hat\yv\com(i)=\fv\com(\x(i))=\boldsymbol{\upphi}\trs(\x(i))\matr{w}(i-1)$.
Therefore, the input-output operation of the composite KLMS algorithm can be expressed as
\begin{align}\LABEQ{compositeLMS}
\hat\yv\com(i)&=\boldsymbol{\upphi}\trs(\x(i))\matr{w}(i-1)\nonumber\\&
=2\mu\sum_{l=1}^{i-1}{\boldsymbol{\upphi}}\trs(\x(i)){\boldsymbol{\upphi}}(\x(l))\matr{e}\com(l)\nonumber\\
&=2\mu\sum_{l=1}^{i-1}\K\left(\x(i),\x(l)\right)\matr{e}\com(l),
\end{align}
where the matrix-valued kernel yields:
\begin{align}\LABEQ{KcompositeLMS}
\K(\x(i),\x(l))&={\boldsymbol{\upphi}}\trs(\x(i)){\boldsymbol{\upphi}}(\x(l))\nonumber\\&=\begin{bmatrix}\PHI\trs\rr(\x(i))\\\PHI\trs\jj(\x(i)\end{bmatrix}\begin{bmatrix}\PHI\rr(\x(l)) \PHI\jj(\x(l))\end{bmatrix}\nonumber\\&=\begin{bmatrix}\PHI\rr\trs(\x(i))\PHI\rr(\x(l)) &{}& \PHI\rr\trs(\x(i))\PHI\jj(\x(l))\\\PHI\jj\trs(\x(i))\PHI\rr(\x(l)) &{}& \PHI\jj\trs(\x(i))\PHI\jj(\x(l))\end{bmatrix}\nonumber\\
&=\begin{bmatrix}\k\rrrr(\x(i),\x(l)) &{}& \k\rrjj(\x(i),\x(l))\\\k\jjrr(\x(i),\x(l)) &{}& \k\jjjj(\x(i),\x(l))\end{bmatrix}.
\end{align}
Notice that this kernel matrix follows the structure introduced in \cite{Boloix17} for the WL-RKHS, and is composed of four scalar real functions.

\section{The proposed generalized complex KLMS algorithm} \LABSEC{AKLM}

Any real-valued composite vector representation $\yv\com=[\yv\trs\rr\; \yv\trs\jj]\trs\in \RN^{2\n}$ of any complex-valued vector $\yv=\yv\rr+\j\yv\jj\in\CN^\n$, can be related to the complex \emph{augmented} vector $\aug\yv=[\yv\trs \;\yv\her]\trs\in\CN^{2\n}$ representation, which is obtained by stacking $\yv$ on top of its complex conjugate $\yv\cnj$. The relation is $\aug{\yv}=\T_\n\yv\com$, where 
\begin{equation} \LABEQ{com2aug}
\T_\n=\left[ \begin{array}{c c}
\I & \j\I \\
\I  & -\j\I \\
\end{array}\right]\in \mathbb{C}^{2\n\times 2\n},
\end{equation}
which is a unitary matrix up to a factor of 2: $\T_\n\T_\n\her=\T_\n\her\T_\n=2\I$, where $\I$ is the identity matrix.

We can now apply this relation to \EQ{compositeLMS} to calculate:
\begin{align}
\hat{\aug{\yv}}(i)&=\begin{bmatrix}\hat\y(i)\\ \hat\y\cnj(i)\end{bmatrix}=\T_1\hat\yv\com(i)
=\T_12\mu\sum_{l=1}^{i-1}\K(\x(i),\x(l))\matr{e}\com(l)\nonumber\\
&=2\mu\sum_{l=1}^{i-1}\T_1\K(\x(i),\x(l))\left(\frac{1}{2}\T_1\her\T_1\right)\matr{e}\com(l)\nonumber\\
&=\mu\sum_{l=1}^{i-1}{\K_A}(\x(i),\x(l))\aug{\matr{e}}(l),\LABEQ{augKLMcompleto}
\end{align}
Here we have the augmented error vector $\aug{\matr{e}}(l)=\T_1\matr{e}\com(l)=[e(l) \; e\cnj(l)]\trs$, and the augmented kernel matrix  
\begin{align}\LABEQ{augkernelmat}
{\K_A}(\x(i),\x(l))&=\T_1\K(\x(i),\x(l))\T_1\her\nonumber\\
&=\begin{bmatrix}
\k(\x(i),\x(l)) & \tilde{\k}(\x(i),\x(l))\\
\tilde{\k}\cnj(\x(i),\x(l)) & \k\cnj(\x(i),\x(l)) 
\end{bmatrix},
\end{align}
where by using \EQ{KcompositeLMS} the complex kernel and complex pseudo-kernel can be identified, respectively, as
\begin{align}
\k(\x(i),\x(l))&=\k\rrrr(\x(i),\x(l))+\k\jjjj(\x(i),\x(l))\nonumber\\&+\j\left(\k\jjrr(\x(i),\x(l))-\k\rrjj(\x(i),\x(l))\right)
\LABEQ{covK},\\
\pk(\x(i),\x(l))&=\k\rrrr(\x(i),\x(l))-\k\jjjj(\x(i),\x(l))\nonumber\\&+\j\left(\k\jjrr(\x(i),\x(l))+\k\rrjj(\x(i),\x(l))\right)
.\LABEQ{pcovK}
\end{align}
Notice that this kernel and pseudo-kernel follow the structure introduced in \cite{Boloix17}.

The first entry of $\hat{\aug{\yv}}(i)$ in \EQ{augKLMcompleto} yields the proposed generalized complex KLMS (gCKLMS):
\begin{align}
\hat{{\y}}(i)=\mu\sum_{l=1}^{i-1}e(l)\k(\x(i),\x(l))+\mu\sum_{l=1}^{i-1}{e}\cnj(l)\pk(\x(i),\x(l)).\LABEQ{augKLMS}
\end{align}


 \section{Connection with other algorithms} \LABSEC{previ}

In \cite{Bouboulis11}, two realizations of the complex-valued KLMS (CKLMS)
algorithm were developed by following two methodologies. The first approach is based on using a complex-valued kernel for a complex RKHS through the associated feature map. In this approach, denoted in \cite{Bouboulis11} as CKLMS2,  the output yields:
\begin{align}
\hat{{\y}}(i)=\mu\sum_{l=1}^{i-1}e(l)\k(\x(i),\x(l)).\LABEQ{CKLMS2}
\end{align}
The second alternative is the \emph{complexification} approach of real RKHSs. In this approach, it is defined the space of complex functions $f(\x)=f_{1}(\x)+\j f_{2}(\x)$ where $f_{1}(\x)$ and $f_{2}(\x)$ are in a RKHS of real functions with real kernel $k_{\Re}$. Then, the complexified real kernel trick allows to construct a kernel adaptive algorithm denoted in \cite{Bouboulis11} as CKLMS1:
\begin{align}
\hat{\y}(i)=\mu\sum_{l=1}^{i-1}2e(l)\k_{\Re}(\x(i),\x(l)).\LABEQ{CKLMS1}
\end{align}
Notice that the kernel used in this CKLMS1 algorithm is a real-valued function.

In \cite{Bouboulis12} it is employed the framework of \cite{Bouboulis11} to develop widely linear adaptive filters in complex RKHS. Two realizations of the augmented CKLMS (ACKLMS) were proposed. First, by using the complexification approach they obtain exactly the same formula \EQ{CKLMS1} for the CKLMS1 algorithm (except for a rescaling) \cite{Bouboulis12}. On the other hand, when a pure complex-valued kernel is used, the ACKLMS algorithm yields
\begin{align}
\hat{{\y}}(i)=\mu\sum_{l=1}^{i-1}\left(e(l)\k(\x(i),\x(l))+e(l)\k\cnj(\x(i),\x(l))\right).\LABEQ{augKLMSBoub1}
\end{align}

At this point it is interesting to note that \EQ{augKLMSBoub1} and \EQ{CKLMS1} are the same. If we take $e(l)$ as a common factor in \EQ{augKLMSBoub1}, it follows:
\begin{align}
\hat{{\y}}(i)&=\mu\sum_{l=1}^{i-1}e(l)\left(\k(\x(i),\x(l))+\k\cnj(\x(i),\x(l))\right)\nonumber\\
&=\mu\sum_{l=1}^{i-1}2e(l)\Re\left\{\k(\x(i),\x({l})\right\}.\LABEQ{augKLMSBoub2}
\end{align}
Hence \EQ{augKLMSBoub1} and \EQ{CKLMS1} provide the same learning process, since in both cases the kernel is real. In fact, they yield the same formula with $\Re\left\{\k\right\}=\k_{\Re}$. 

Next, we show that algorithms CKLMS1, CKLMS2 \cite{Bouboulis11} and ACKLMS \cite{Bouboulis12} are particular cases of our proposed gCKLMS algorithm in \EQ{augKLMS}. They yield a subset of the cases the gCKLMS algorithm presented in this paper can represent. 

First, these approaches do not have a pseudo-kernel term, therefore they provide simplified limited versions and hence a reduction on the flexibility the general algorithm provides. It is easy to check that if we set the pseudo-kernel equal to zero in \EQ{augKLMS} the gCKLMS reduces to the CKLMS2 in \EQ{CKLMS2}. However, to have $\pk(\x(i),\x(l))=0$ in \EQ{pcovK} the following conditions must be satisfied:
\begin{align}
\k\rrrr(\x(i),\x(l))&=\k\jjjj(\x(i),\x(l)),\nonumber\\
\k\jjrr(\x(i),\x(l))&=-\k\rrjj(\x(i),\x(l))),\LABEQ{condiciones}
\end{align}
and the kernel in \EQ{covK} yields  
\begin{align}
\k(\x(i),\x(l))=2\k\rrrr(\x(i),\x(l))-\j2\k\rrjj(\x(i),\x(l)).\LABEQ{kernelsimple}
\end{align} 

Second, if in addition to $\pk(\x(i),\x(l))=0$ we now set $\k\rrjj(\x(i),\x(l))=0$, then the kernel in \EQ{kernelsimple} becomes a real-valued function $\k(\x(i),\x(l))=2\k\rrrr(\x(i),\x(l))$, and the gCKLMS simplifies to the CKLMS1 in \EQ{CKLMS1} or the ACKLMS in \EQ{augKLMSBoub2}. 

In Table \ref{table:relac} we summarize the algorithms and the conditions they impose on the kernel and pseudo-kernel terms.

\begin{table*}[bth]
\caption{Conditions imposed on the kernel and pseudo-kernel by the algorithms}
\label{table:relac}
\centering
\begin{tabular}{p{0.15\linewidth}p{0.24\linewidth}p{0.24\linewidth}p{0.24\linewidth}} 
\hline
Algorithm & Kernel term& Pseudo-kernel term & Conditions\\
\hline
gCKLMS & $\k{}=\k\rrrr{}+\k\jjjj{}\nonumber+\j\left(\k\jjrr{}-\k\rrjj{}\right)$ in \EQ{covK} & $\pk{}=\k\rrrr{}-\k\jjjj{}\nonumber+\j\left(\k\jjrr{}+\k\rrjj{}\right)$ in \EQ{pcovK} & {}\\
CKLMS2 & $\k{}=2\k\rrrr{}-\j2\k\rrjj{}$ & $\pk{}=0$ & $\k\rrrr=\k\jjjj$, $\k\jjrr=-\k\rrjj$ in \EQ{condiciones}\\
ACKLMS/CKLMS1 & $\k{}=2\k\rrrr{}\in\RN$ & $\pk{}=0$ & $\k\rrrr=\k\jjjj$, $\k\jjrr=-\k\rrjj=0$\\
\hline
\end{tabular}
\end{table*}


 \subsection{Kernel design} \LABSEC{kerneldesign}
 
The conditions that the algorithms impose on the kernel and pseudo-kernel terms must be carefully analyzed in order to choose the best algorithm and kernels for a given learning problem. 

The kernel in a RKHS learning algorithm encodes our assumptions about the function that is being learned \cite{Scholkopf02} and provides a measure of similarity  between the inputs. In \cite{Boloix17} the kernel and pseudo-kernel in \EQ{covK}-\EQ{pcovK} are analyzed, and several remarks are provided to help designing them and deciding when they should be real or complex-valued. We use that analysis here to understand the implications of the conditions that each algorithm impose.

We start with the conditions imposed when the pseudo-kernel is null, i.e., the conditions in \EQ{condiciones} that yield the complex-valued kernel in \EQ{kernelsimple}. 
For any two inputs $\x$ and $\x'$, the first condition $\k\rrrr(\x,\x')=\k\jjjj(\x,\x')$ implies that the same measure of similarity must be used with the real and the imaginary parts of the function \cite{Boloix17}. Hence, if we impose a null pseudo-kernel, we cannot use a kernel for the real part, {$\k\rrrr(\x,\x')$}, and another different design for the imaginary part, {$\k\jjjj(\x,\x')$}. The second condition is $\k\jjrr(\x,\x')=-\k\rrjj(\x,\x')$. But we also have $\k\jjrr(\x,\x')=\k\rrjj(\x,\x')$, because the kernel matrix $\K(\x,\x')$ in \EQ{KcompositeLMS} is positive semi-definite. Therefore, $\k\rrjj(\x,\x')=-\k\rrjj(\x,\x')$. This imposes a skew-symmetry in the measure of similarity between the real and the imaginary parts of the function.

As an example, the complex Gaussian kernel proposed in \cite{Bouboulis11} for the CKLMS2 algorithm:
\begin{align} \LABEQ{compGaussKernel}
k_{\CN G}(\x,\x')&=\exp\left(-(\x-\x'^*)^\top(\x-\x'^*)/\gamma_{\CN G}^2\right),
\end{align}
follows the form given in \EQ{kernelsimple} and fulfils the conditions in \EQ{condiciones}, i.e., the symmetries that yield a null pseudo-kernel. However, this kernel measures similarities between the real parts of the inputs with $|\x\rr-\x\rr'|^2$, while for the imaginary ones it uses $|\x\jj+\x\jj'|^2$, where $|\cdot|$ is the $\ell^2$-norm. Also, it is not stationary, has an oscillatory behavior, and the exponent in the kernel may easily grow large and positive \cite{Boloix15}. This might cause numerical problems and, as we show later in the experiments, it does not yield the best performance.

The skew-symmetry $\k\rrjj(\x,\x')=-\k\rrjj(\x,\x')$ imposed for the CKLMS2 algorithm may not be a property satisfied by many to-be-learned functions and, in such a case, enforcing a complex-valued kernel can be counterproductive. Algorithms CKLMS1 \cite{Bouboulis11} and ACKLMS \cite{Bouboulis12} avoid this problem by adding another condition: $\k\rrjj(\x(i),\x(l))=0$. Therefore, these algorithms use a real-valued kernel $\k(\x,\x')=2\k\rrrr(\x,\x')$. The condition $\k\rrjj(\x(i),\x(l))=0$ implies that the real and the imaginary parts are not related and that one of them does not provide information to learn the other \cite{Boloix17}.  

We conclude that algorithms CKLMS1, CKLMS2 \cite{Bouboulis11} and ACKLMS \cite{Bouboulis12} cannot represent any possible complex-valued function, and yield a subset of the cases that the gCKLMS algorithm proposed in this paper can represent. The gCKLMS, with the kernel and the pseudo-kernel terms, provides more flexibility to model the learning problem by means of the four real-valued functions $\k\rrrr$, $\k\jjjj$, $\k\rrjj$ and $\k\jjrr$. Hence, the gCKLMS will provide the best result if the conditions described above are not suitable for our learning problem, i.e., when the real and imaginary parts are better represented with different kernels, or they are not independent, or the skew-symmetry imposed by $\k\rrjj(\x,\x')=-\k\rrjj(\x,\x')$ does not hold.

We end this discussion about the kernels by bringing here a suitable real-valued function for $\k\rrrr$, $\k\jjjj$, $\k\rrjj$ and $\k\jjrr$ proposed in \cite{Boloix17}. This is the adaptation to complex-valued inputs of the real-valued Gaussian kernel: 
\begin{equation} \LABEQ{expkernel}
\k_{G}(\x,\x')=\exp\left(-(\x-\x')\her(\x-\x')/\gamma^2\right),
\end{equation}
where $\gamma$ is the kernel parameter. This real function provides a measure of similarity between the complex-valued inputs that is simple but effective for complex-valued signals: inputs closer to other input in the complex field are considered more similar than inputs that are further away \cite{Boloix17}. We will use it in our experiments.
For a further analysis about the selection of suitable kernels for complex-valued applications see \cite{Boloix17, Boloix15}.


\section{Experiments}\LABSEC{Exp}

We consider two experiments where we compare the performance of our proposal the gCKLMS in \EQ{augKLMS}, versus the CKLMS2 in \EQ{CKLMS2} \cite{Bouboulis11} and the ACKLMS algorithm in \EQ{augKLMSBoub2} \cite{Bouboulis12}.

 In the first experiment we reproduce the nonlinear channel equalization task in \cite{Bouboulis12}. In this experiment the complex-valued signals have independent real and imaginary parts, and they are better represented with different kernels. We show that in such a case the best choice is a real kernel and a real pseudo-kernel. 
 
 In the second experiment we propose learning a filtered two-dimensional random process. At the output of the filter the real and imaginary parts are not independent, and we show that we can use the imaginary part of the pseudo-kernel to improve the performance.

As in \cite{Bouboulis12}, we use the complex Gaussian kernel $k_{\CN G}(\x,\x')$ in \EQ{compGaussKernel} for both the CKLMS2 and the ACKLMS. In fact, for the ACKLMS the real part of this kernel is used, as was shown in \EQ{augKLMSBoub2}. We use the code available in \cite{boubouliscode} to run the algorithms.

For our proposed gCKLMS we use the general kernel and pseudo-kernel in \EQ{covK}-\EQ{pcovK}. For $\k\rrrr(\x,\x')$, $\k\jjjj(\x,\x')$, $\k\jjrr(\x,\x')$ and $\k\rrjj(\x,\x')$ we propose to use the real-valued Gaussian kernel $\k_{G}(\x,\x')$ in \EQ{expkernel} with parameter $\gamma=\gamma\rr$ for $\k\rrrr$, $\gamma=\gamma\jj$ for $\k\jjjj$, $\gamma=\gamma\rrjj$ for $\k\rrjj$, and $\gamma=\gamma\jjrr$ for $\k\jjrr$, respectively. The kernel and pseudo-kernel can be simplified if the signals meet any of the conditions discussed in \SEC{kerneldesign}. For example, if the real and imaginary parts of the signals are independent we can set $\k\jjrr(\x,\x')=\k\rrjj(\x,\x')=0$ and the kernel and pseudo-kernel are real-valued:
\begin{align} 
\k(\x,\x')&=\k\rrrr(\x,\x')+\k\jjjj(\x,\x'),\LABEQ{realk}\\
\pk(\x,\x')&=\k\rrrr(\x,\x')-\k\jjjj(\x,\x').\LABEQ{realpk}
\end{align}
We use this simplification in the first experiment.
Notice that if we also assume that the real and imaginary parts of the output use the same kernel, $\k\rrrr=\k\jjjj$, then we should set $\gamma\rr=\gamma\jj$ and the pseudo-kernel term in \EQ{realpk} cancels. In such a case, as explained in \SEC{previ}, the gCKLMS approach simplifies to the ACKLMS with real-valued kernel $\k(\x,\x')=2\k\rrrr(\x,\x')$, where $\k\rrrr(\x,\x')$ is as in \EQ{expkernel} with $\gamma=\gamma\rr$. We will refer to this case as ACKLMS with kernel \EQ{expkernel} in the experiments.

\begin{figure}[t!b]
\begin{center}
\includegraphics[width=8.6cm, draft=false]{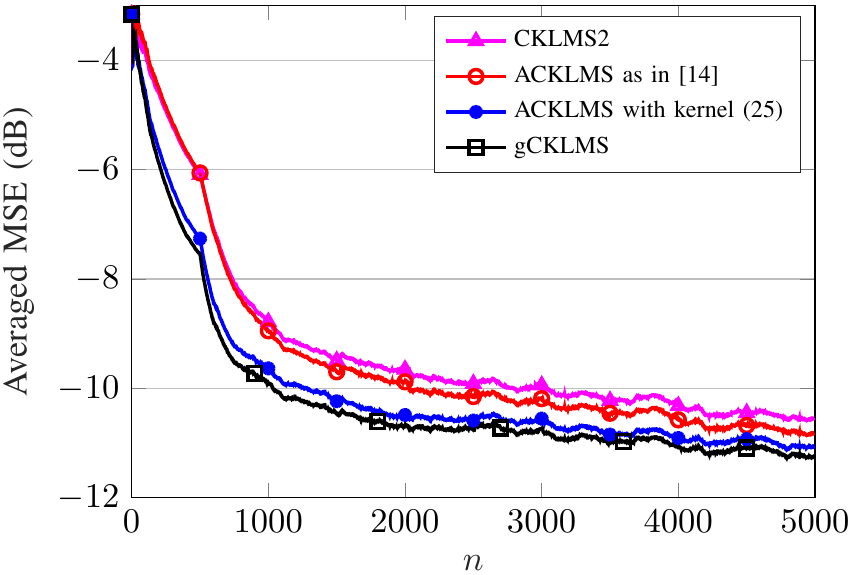}
\end{center}
\vspace*{-.6cm}
\caption{MSE in dB versus the number of input samples for the \emph{soft nonlinear channel} with the circular input case for the CKLMS2 ($\gamma_{\CN G}=10$), the ACKLMS as in \cite{Bouboulis12} ($\gamma_{\CN G}=10$), the ACKLMS with kernel \EQ{expkernel} ($\gamma\rr =5$), and the gCKLMS ($\gamma\rr=6.5$ and $\gamma\jj=5.5$).}
\LABFIG{figBoub1}
\end{figure}
\begin{figure}[t!b]
\begin{center}
\includegraphics[width=8.6cm, draft=false]{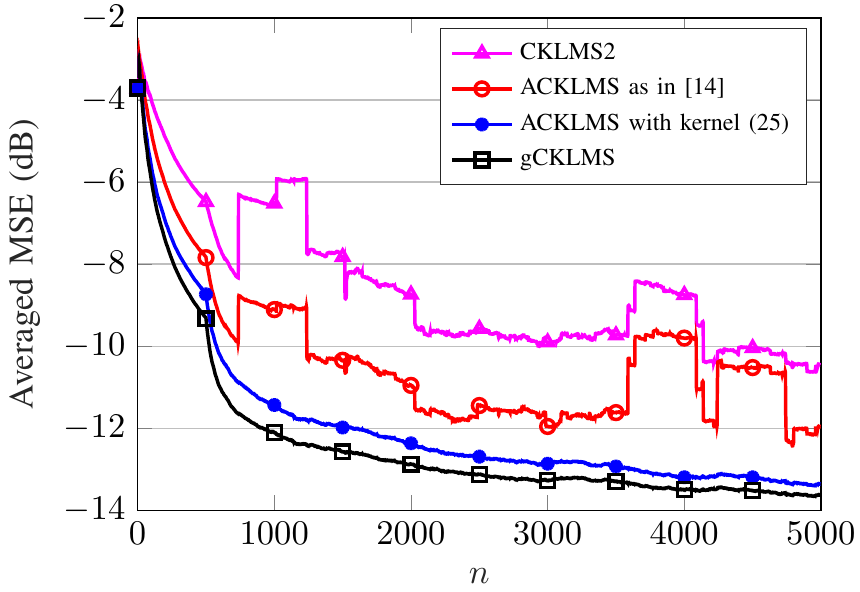}
\end{center}
\vspace*{-.6cm}
\caption{MSE in dB versus the number of input samples for the \emph{soft nonlinear channel} with the noncircular input case for the CKLMS2 ($\gamma_{\CN G}=10$), the ACKLMS as in \cite{Bouboulis12} ($\gamma_{\CN G}=10$), the ACKLMS with kernel \EQ{expkernel} ($\gamma\rr =5$), and the gCKLMS ($\gamma\rr=6.5$ and $\gamma\jj=5.5$).}
\LABFIG{figBoub2}
\end{figure}

\subsection{Nonlinear channel equalization}

We face the problem of nonlinear channel equalization, as in \cite{Bouboulis11} and \cite{Bouboulis12}, for ease of comparison and continuity.
The channel consists of a linear filter and a memoryless nonlinearity. The two nonlinear channels in \cite{Bouboulis12} have been considered here. The
first channel is the \emph{soft nonlinear channel}, with linear filter 
\begin{align}
t(n)=(-0.9+0.8\j)\cdot s(n)+(0.6-0.7\j)\cdot s(n-1),\nonumber
\end{align}
followed by the nonlinearity 
\begin{align}
q(n)=t(n)+(0.1+0.15\j)\cdot t^{2}(n)+ (0.06+0.05\j)\cdot t^{3}(n).\nonumber
\end{align}
The second one is the \emph{strong nonlinear channel}, with linear filter
\begin{align}
t(n)=&(-0.9+0.8\j)\cdot s(n)+(0.6-0.7\j)\cdot s(n-1)\nonumber\\&+(-0.4+0.3\j)\cdot s(n-2)+(0.3-0.2\j)\cdot s(n-3)\nonumber\\&+(-0.1-0.2\j)\cdot s(n-4),\nonumber
\end{align}
and nonlinearity 
\begin{align}
q(n)=t(n)+(0.2+0.25\j)\cdot t^{2}(n)+ (0.08+0.09\j)\cdot t^{3}(n).\nonumber
\end{align}

\begin{figure}[t!b]
\begin{center}
\includegraphics[width=8.6cm, draft=false]{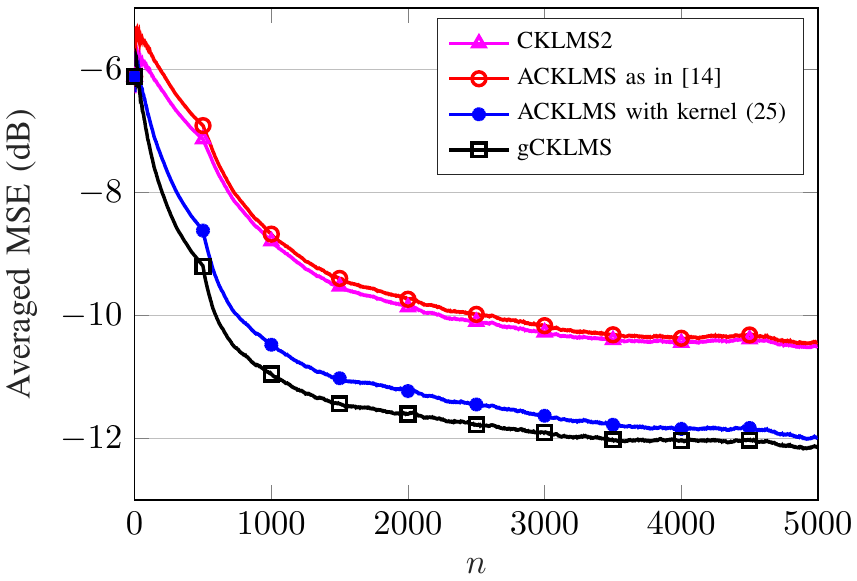}
\end{center}
\vspace*{-.6cm}
\caption{MSE in dB versus the number of input samples for the \emph{strong nonlinear channel} with the circular input case for the CKLMS2 ($\gamma_{\CN G}=15$), the ACKLMS as in \cite{Bouboulis12} ($\gamma_{\CN G}=15$), the ACKLMS with kernel \EQ{expkernel} ($\gamma\rr =5$), and the gCKLMS ($\gamma\rr=5$ and $\gamma\jj=3$).}
\LABFIG{figBoub3}
\end{figure}
\begin{figure}[t!b]
\begin{center}
\includegraphics[width=8.6cm, draft=false]{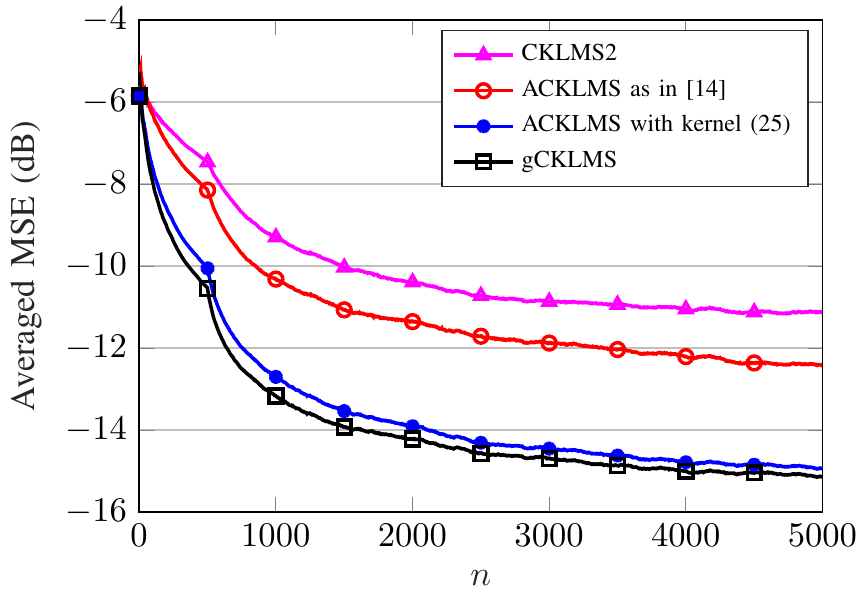}
\end{center}
\vspace*{-.6cm}
\caption{MSE in dB versus the number of input samples for the \emph{strong nonlinear channel} with the noncircular input case for the CKLMS2 ($\gamma_{\CN G}=15$), the ACKLMS as in \cite{Bouboulis12} ($\gamma_{\CN G}=15$), the ACKLMS with kernel \EQ{expkernel} ($\gamma\rr=5$), and the gCKLMS ($\gamma\rr=5$ and $\gamma\jj=3$).}
\LABFIG{figBoub4}
\end{figure}

At the receiver, the signal $q(n)$ is corrupted by additive white circular Gaussian noise with an SNR of 15 dB to yield the received signal $r(n)$. The inputs to the equalizer are the sets of samples $\vect{x}=[r(n+D),r(n+D-1),\cdots,r(n+D-L+1)]^{\top}$, where $L>0$ is the filter length and $D$ is the equalization time delay. Here we set $L=5$ and $D=2$, as in \cite{Bouboulis12}. The goal is to estimate the
original input signal $s(n)$. 
\newline\vspace{-3mm}

\subsubsection{Gaussian distributed inputs}

We first set the input signals as in \cite{Bouboulis12}: $s(n)=0.7(\sqrt{1-\rho^2}\cdot X(n)+\j \rho\cdot Y(n))$, where $X(n)$ and $Y(n)$ are independent Gaussian random variables, with $\rho=1/\sqrt{2}$ for circular signals, and $\rho=0.1$ for noncircular signals. 

The real and imaginary parts of the signals are independent and, therefore, we can set $\k\jjrr(\x,\x')=\k\rrjj(\x,\x')=0$ and use the real-valued kernel and pseudo-kernel terms in \EQ{realk}-\EQ{realpk} for our proposed gCKLMS. 

Experiments were conducted on 100 independent sets of 5000 samples of the
input signal. 
For all the approaches, the novelty criterion \cite{Liu10,Platt91}, is used for sparsification with $\delta_{1}=0.15$ and $\delta_{2}=0.2$, as in \cite{Bouboulis12}.

Figs. \ref{fig:figBoub1} and \ref{fig:figBoub2} show the averaged mean square errors (MSE) for the \emph{soft nonlinear channel}. The circular input case is shown in \FIG{figBoub1}, and the noncircular input case is shown in \FIG{figBoub2}. For the CKLMS2 and the ACKLMS algorithms we set $\gamma_{\CN G}=10$ and $\mu=1/8$, as in \cite{Bouboulis12}. For the gCKLMS algorithm we set $\gamma\rr=6.5$ and $\gamma\jj=5.5$, and $\mu=1/7$. For the ACKLMS with kernel \EQ{expkernel} we set $\gamma\rr=5$ and $\mu=1/10$. 

Figs. \ref{fig:figBoub3} and \ref{fig:figBoub4} include the MSE for the \emph{strong nonlinear channel} and the circular input  and noncircular input cases, respectively. 
For the CKLMS2 and the ACKLMS algorithms $\gamma_{\CN G}=15$ and $\mu=1/6$, as in \cite{Bouboulis12}. For the gCKLMS algorithm we set $\gamma\rr=5$ and $\gamma\jj=3$, and $\mu=1/7$. For the ACKLMS with kernel \EQ{expkernel} we set $\gamma\rr=5$ and $\mu=1/10$.

In all the examples, the proposed gCKLMS outperforms the other algorithms. 
The main advantage of the gCKLMS is that by introducing a pseudo-kernel we can use a different kernel for the real and the imaginary parts, $\k\rrrr(\x,\x')$ and $\k\jjjj(\x,\x')$.
This extra degree of freedom, which is not present in the other algorithms, is the key to obtain a better estimation. In this experiment, the gain in MSE is small, because $\k\rrrr(\x,\x')$ and $\k\jjjj(\x,\x')$ are very similar. That is the reason why the ACKLMS with kernel \EQ{expkernel}, i.e., setting $\gamma\rr=\gamma\jj$, performs well and close to the general case. In any case, it can be observed that to achieve a given error, the faster convergence of the gCKLMS allows saving 10\%-30\% of the samples and time. 

With the complex Gaussian kernel, both the ACKLMS in \cite{Bouboulis12} and the CKLM2  in \cite{Bouboulis11} perform poorly compared to the gCKLMS. Therefore, the experiments show that the complex Gaussian kernel is not the best choice for this equalization task and, as it is shown in \FIG{figBoub2}, sometimes yields undesired spikes in the learning curves.
\newline\vspace{-2mm}

\begin{figure}[!tb]
\begin{center}
\includegraphics[width=8.6cm, draft=false]{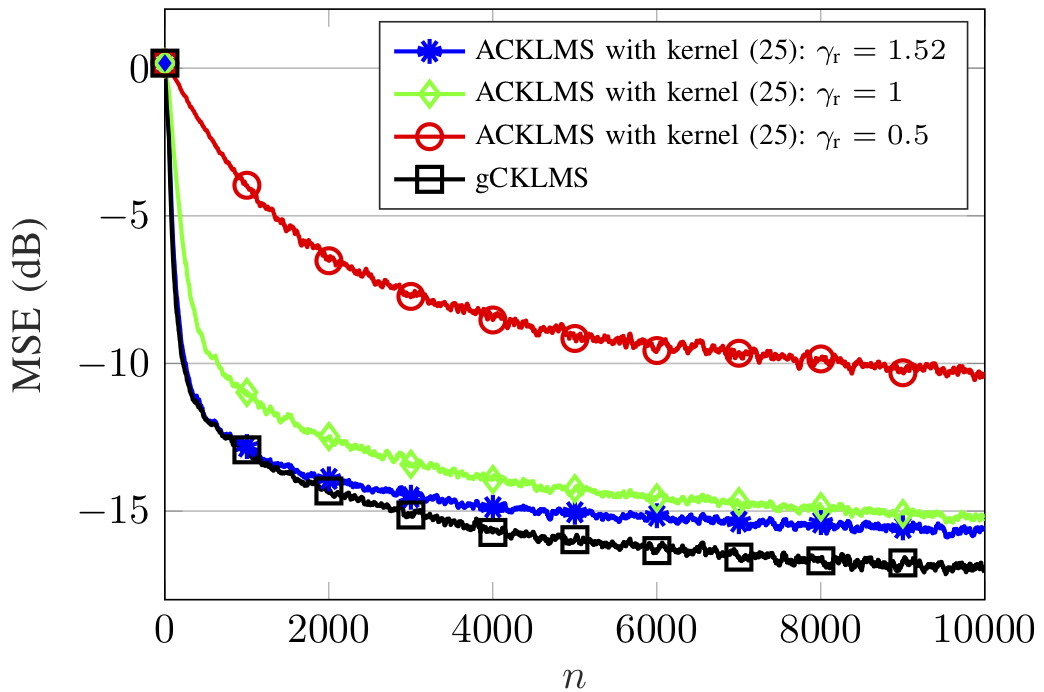}
\end{center}
\vspace*{-.6cm}
\caption{MSE in dB versus the number of input samples for the ACKLMS with kernel \EQ{expkernel} ($\gamma\rr = \{0.5, 1, 1.52\}$) and the gCKLMS ($\gamma\rr=0.59$, $\gamma\jj=1.63$).}
\LABFIG{fig4}
\end{figure}

\begin{figure}[!tb]
\begin{center}
\includegraphics[width=8.6cm, draft=false]{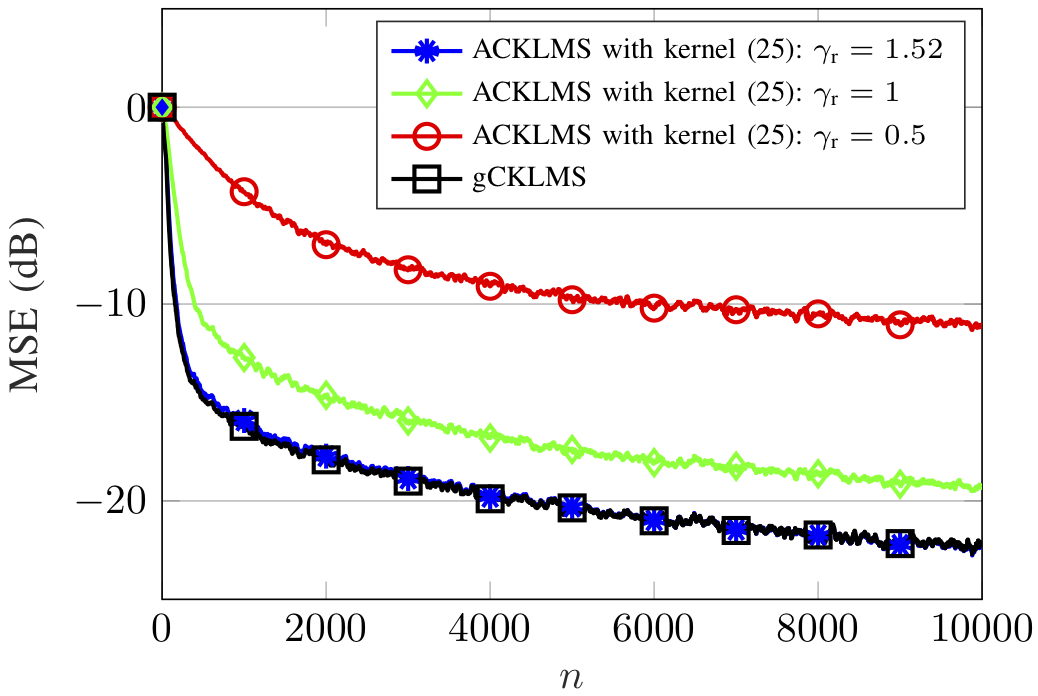}
\end{center}
\vspace*{-.6cm}
\caption{MSE in dB for the imaginary part versus the number of input samples for the ACKLMS with kernel \EQ{expkernel} ($\gamma\rr = \{0.5, 1, 1.52\}$) and the gCKLMS ($\gamma\rr=0.59$, $\gamma\jj=1.63$).}
\LABFIG{fig5}
\end{figure}

\begin{figure}[!hbt]
\begin{center}
\includegraphics[width=8.6cm, draft=false]{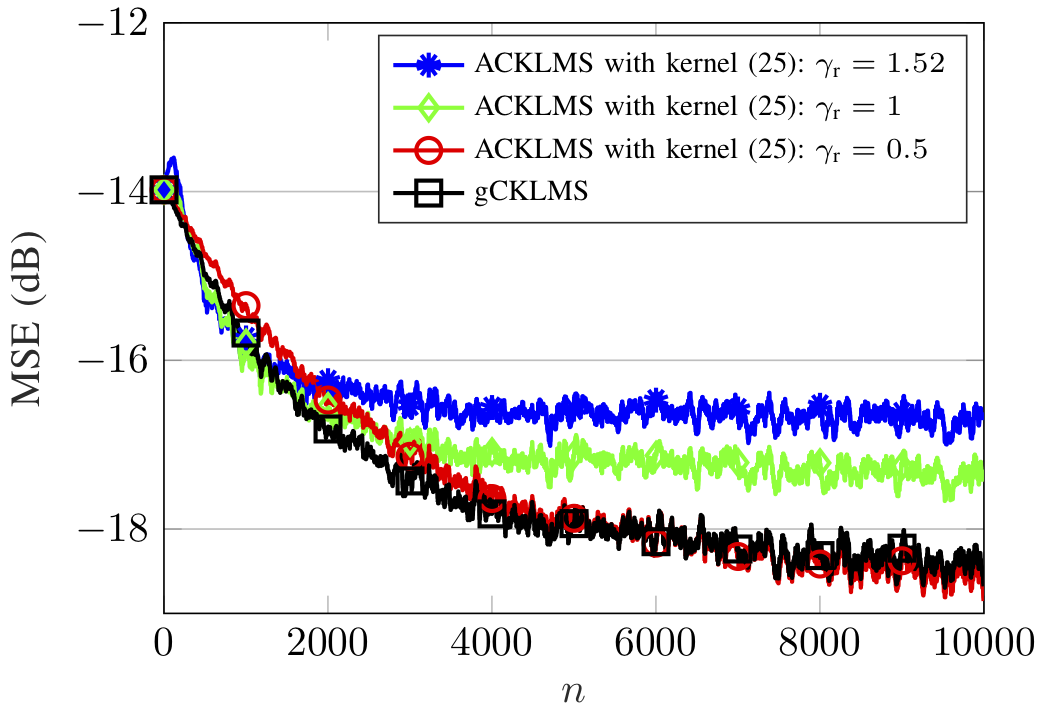}
\end{center}
\vspace*{-.6cm}
\caption{MSE in dB for the real part versus the number of input samples for the ACKLMS with kernel \EQ{expkernel} ($\gamma\rr = \{0.5, 1, 1.52\}$) and the gCKLMS ($\gamma\rr=0.59$, $\gamma\jj=1.63$).}
\LABFIG{fig6}
\end{figure}

\subsubsection{Unbalanced digital modulated signals}
In digital communications inputs are discrete. For discrete and unbalanced digital modulated signals, 
the difference between $\k\rrrr(\x,\x')$ and $\k\jjjj(\x,\x')$ is greater and the proposed gCKLMS algorithm is a good choice versus the previous proposals, with null pseudo-kernels. 
To illustrate this point, we propose to repeat here the equalization experiment for the \emph{soft nonlinear channel}, 
where the input signals are now $s(n)=0.2 X(n)+\j 0.1Y(n)$, where $X(n)$ and $Y(n)$ are independent binary $\{-1,+1\}$ data streams. 

For the proposed gCKLMS algorithm we use again the real-valued Gaussian kernel \EQ{expkernel} with parameters $\gamma\rr=0.59$ for $\k\rrrr(\x,\x')$ and $\gamma\jj=1.63$ for $\k\jjjj(\x,\x')$, respectively. For the ACKLMS with kernel \EQ{expkernel} we set $\gamma\rr=1.52$. 
The learning parameter is set to $\mu=0.5$ for both approaches. 

We generate 100 independent test trials with a set of 10000 samples to test the algorithms. The mean square errors (MSE) of the estimation are compared in \FIG{fig4} versus the number of input samples. It can be observed that the proposed gCKLMS outperforms the ACKLMS with kernel \EQ{expkernel}, i.e., the case with null pseudo-kernel.

Again, the key to obtain a better estimation with the gCKLMS in this experiment is the possibility to define a different kernel for the real part $\k\rrrr(\x,\x')$ and the imaginary part $\k\jjjj(\x,\x')$. Figs. \ref{fig:fig5} and \ref{fig:fig6} are included to highlight this. Fig. \ref{fig:fig5} shows the estimation MSE of the imaginary part of the signals, while Fig. \ref{fig:fig6} shows the estimation MSE of the real part. In this experiment the real and imaginary parts require a different kernel to be accurately learnt. However, the ACKLMS algorithm uses the same kernel for both parts. The parameter value $\gamma\rr$ with best performance to learn the imaginary part of the output in Fig. \ref{fig:fig5} yields the worst estimation of the real part in Fig. \ref{fig:fig6}. And vice versa, the best parameter value to learn the real part of the output in Fig. \ref{fig:fig6} provides the worst performance in Fig. \ref{fig:fig5}. 
%
%
Remarkably, the estimation with the proposed gCKLMS is always low for both imaginary and real parts, as it allows to set different values for $\k\rrrr(\x,\x')$ and $\k\jjjj(\x,\x')$.


\begin{figure}[!hbt]
\begin{center}
\includegraphics[width=8.6cm, draft=false]{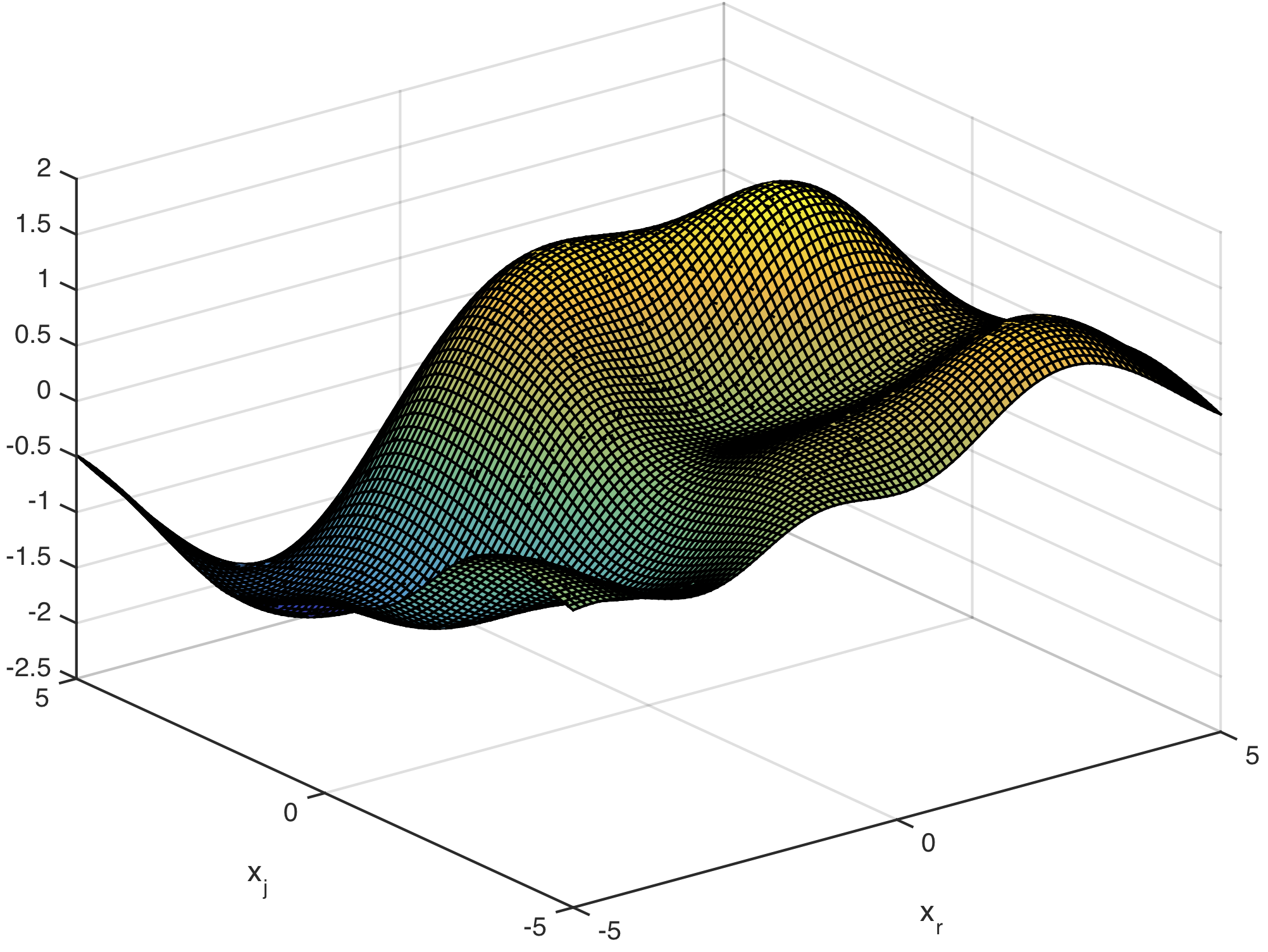}
\end{center}
\vspace*{-.6cm}
\caption{Real part of a sample of the filtered process.}
\LABFIG{fig1}
\end{figure}
\begin{figure}[!hbt]
\begin{center}
\includegraphics[width=8.6cm, draft=false]{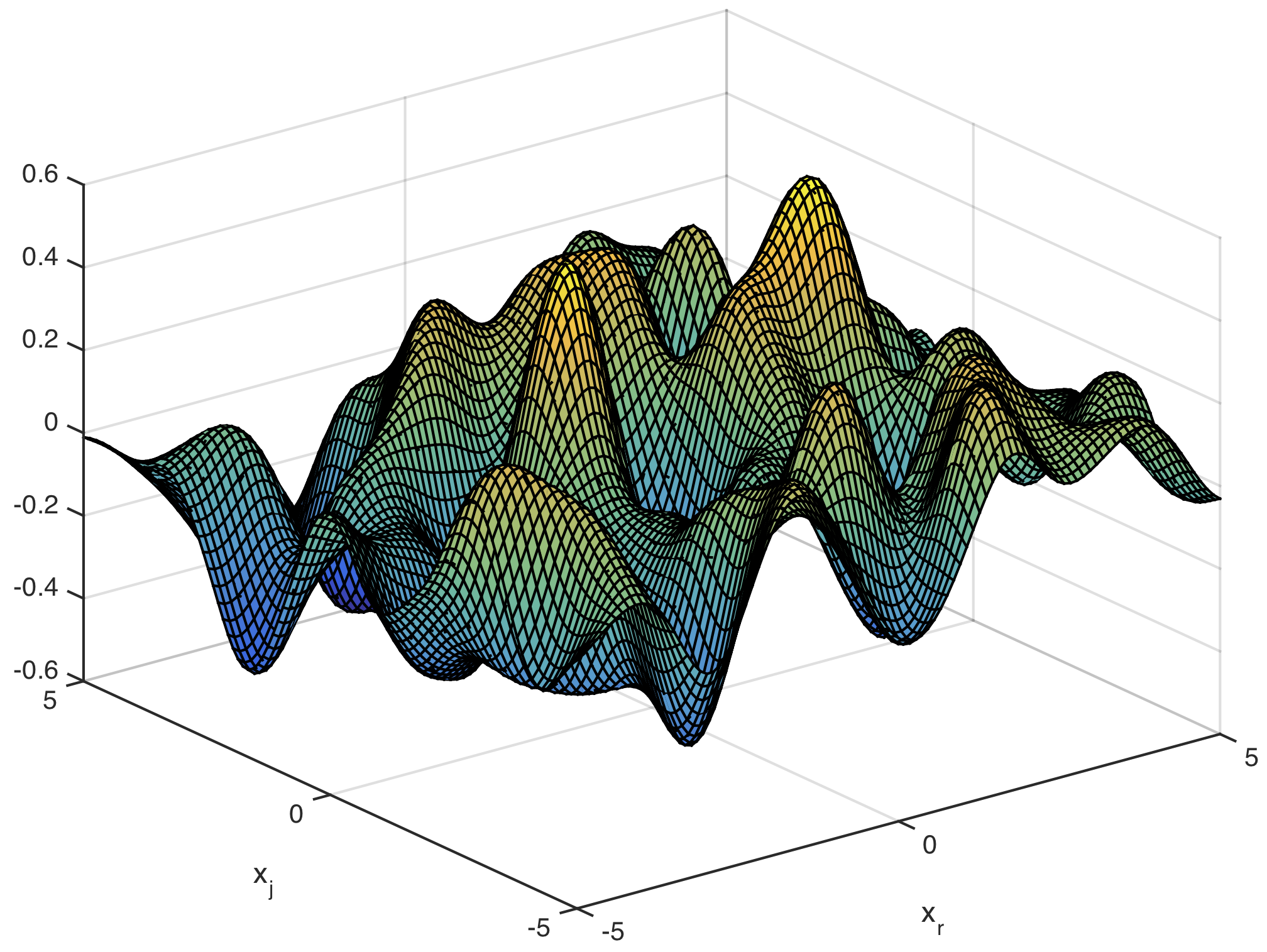}
\end{center}
\vspace*{-.6cm}
\caption{Imaginary part of a sample of the filtered process.}
\LABFIG{fig2}
\end{figure}
\begin{figure}[!hbt]
\begin{center}
\includegraphics[width=8.6cm, draft=false]{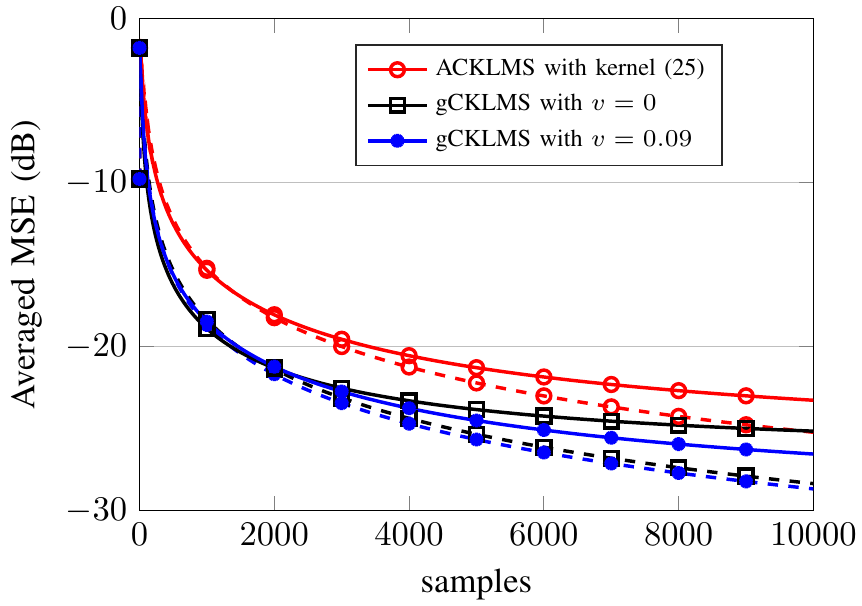}
\end{center}
\vspace*{-.6cm}
\caption{Averaged MSE in dB versus the number of input samples for the ACKLMS ($\gamma\rr=0.76$), the  gCKLMS with $v=0$ ($\gamma\rr=1.62$ and $\gamma\jj=0.59$), and the  gCKLMS with $v=0.09$ ($\gamma\rr=1.73$, $\gamma\jj=0.58$, $\gamma\jjrr=1.11$). Dashed lines: SNR = 50 dB. Solid lines: SNR = 15 dB.} \LABFIG{fig_rand}
\end{figure}

\subsection{A random process filtered}
In this experiment, we show the performance of the proposed gCKLMS when the signals do not have independent real and imaginary parts. We generate a complex-valued signal with correlated real and imaginary parts by filtering a real valued random process with a complex-valued filter. We define the complex-valued filter $h(x)=\alpha\cdot\left(2\exp(-|x|^2/3)+\j\exp(-|x|^2/0.5\right)$, where $x=x\rr+\j x\jj$, with $x\rr\in[-5,5]$ and $x\jj\in[-5,5]$, and $\alpha=0.0228$ to ensure unit norm. Then we define a real Gaussian process $s(x\rr,x\jj)$ with zero mean and unit variance, and we pass this process through the filter. We show in Figs. \ref{fig:fig1} and \ref{fig:fig2} the real and imaginary parts of one sample of the filtered process in $x\rr\in[-5,5]$ and $x\jj\in[-5,5]$. We adaptively learn this filtered process with our proposed gCKLMS algorithm and compare its performance with the ACKLMS with kernel \EQ{expkernel}. 

For the gCKLMS, we make use of the terms $\k\jjrr(\x,\x')$ and $\k\rrjj(\x,\x')$ in \EQ{covK}-\EQ{pcovK}. Since there is a relationship between the real and the imaginary parts of the filtered process, we do not assume these terms to be zero in this experiment,  but we assume that $\k\jjrr(\x,\x')=\k\rrjj(\x,\x')$. Hence, \EQ{covK}-\EQ{pcovK} yield:
\begin{align} 
\k(\x,\x')&=\k\rrrr(\x,\x')+\k\jjjj(\x,\x'),\\
\pk(\x,\x')&=\k\rrrr(\x,\x')-\k\jjjj(\x,\x')+2\j\k\jjrr(\x,\x'),\LABEQ{compk}
\end{align}
and we use a real-valued kernel and a complex-valued pseudo-kernel \cite{Boloix17}. For $\k\rrrr(\x,\x')$ and $\k\jjjj(\x,\x')$ we again use the real-valued Gaussian kernel $\k_{G}(\x,\x')$ in \EQ{expkernel} with parameters $\gamma\rr=1.73$ and $\gamma\jj=0.58$. For the imaginary part of the pseudo-kernel, we use $\k\jjrr(\x,\x')=v\cdot\k_{G}(\x',\x')$ with parameter $\gamma\jjrr=1.11$. Note that the variable $v$ controls the amplitude of the imaginary part of the pseudo-kernel. It was set to $v=0.09$. The learning step was set to $\mu=1/4$.

For the ACKLMS we again propose $\k(\x,\x')=2\k\rrrr(\x,\x')$, with $\k\rrrr(\x,\x')$ the real-valued Gaussian kernel $\k_{G}(\x,\x')$ in \EQ{expkernel}, since it yields better performance than the complex Gaussian kernel. The kernel parameter is set to $\gamma\rr=0.76$ and $\mu=1/2$.

We generate $100$ independent samples of the filtered process in $x\rr\in[-5,5]$ and $x\jj\in[-5,5]$, each with $10000$ data points. A random white circular Gaussian noise is added to the samples. The averaged MSE of the estimation are compared in \FIG{fig_rand} versus the number of input points for two values of SNR, $15$ and $50$ dB. The proposed gCKLMS algorithm greatly outperforms the ACKLMS algorithm. We also include in the figure the performance of the gCKLMS when $v=0$, i.e., when both the kernel and pseudo-kernel are real-valued. In such a case, the kernel parameters have been slightly modified to $\gamma\rr=1.62$ and $\gamma\jj=0.59$ in order to reach the best result. The figure shows that the imaginary part of the pseudo-kernel helps to improve the prediction accuracy by making use of the $\k\jjrr(\x,\x')$ term in the pseudo-kernel.

\section{Conclusions} \LABSEC{Conclu}

In this paper, we have developed a novel generalized formulation for the complex-valued KLMS algorithm. Based on the ideas recently presented in \cite{Boloix17} for the WL-RKHS, we have developed the gCKLMS algorithm that includes both a kernel and a pseudo-kernel. We reviewed the theory of RKHS of  vector-valued functions to define the feature map for the RKHS of composite vector-valued functions. Based on this definition, we were able to develop the composite KLMS algorithm to later rewrite it in augmented notation and, finally, yield the proposed gCKLMS algorithm. Also, in this process we were able to identify the equations that define the kernel and pseudo-kernel. These equations follow the structure introduced in \cite{Boloix17}, and include four real-valued functions: $\k\rrrr(\x,\x')$, $\k\jjjj(\x,\x')$, $\k\rrjj(\x,\x')$ and $\k\jjrr(\x,\x')$. We can use the analysis in \cite{Boloix17} to design this real-valued functions and set the kernel and pseudo-kernel for a given application. Another important contribution of the paper is to show that previous proposed complex-valued KLMS algorithms are just particular simplifications of the gCKLMS proposed in this paper. The experiments included reveal that the gain of using the gCKLMS algorithm, which provides more flexibility than the previous proposed algorithms, can be significant.


%


\ifCLASSOPTIONcaptionsoff
  \newpage
\fi



%

\bibliographystyle{IEEEtran}
\bibliography{IEEEabrv,CGPR,murilloGP,SSCDMA}

%
%

%








\end{document}